\documentclass[runningheads]{llncs}
\usepackage[T1]{fontenc}
\usepackage{graphicx}
\usepackage{times}
\usepackage{epsfig}
\usepackage{amsmath}
\usepackage{amssymb}
\usepackage{booktabs}
\usepackage{subcaption}
\usepackage{bm}
\usepackage{wrapfig}

\begin{document}
\title{An Experimental Investigation into the Evaluation of Explainability Methods}
\titlerunning{An Expr. Invest. into the Eval. of XAI Methods}
\author{Sédrick Stassin\inst{1} \and
Alexandre Englebert\inst{2,3} \and
Géraldin Nanfack\inst{4} \and Julien Albert\inst{4} \and Nassim Versbraegen\inst{5,6} \and Gilles Peiffer\inst{2} \and Miriam Doh\inst{8,9} \and Nicolas Riche\inst{7} \and Benoît Frenay\inst{4} \and Christophe De Vleeschouwer\inst{2}}
\authorrunning{S. Stassin et al.}
%

\institute{ILIA unit, Faculty of Engineering, University of Mons, Mons, Belgium \and
Information and Communication Technologies, Electronics and Applied Mathematics, Louvain-la-Neuve, UCLouvain, Belgium \and
Service de chirurgie orthopédique et traumatologie, Cliniques Universitaires Saint-Luc UCL
\& Neuro Musculo Skeletal Lab (NMSK), IREC, Woluwe-Saint-Lambert, UCLouvain, Belgium \and
PReCISE Research Center, Namur Digital Institute (NADI), University of Namur, Belgium \and
Interuniversity Institute for Bioinformatics in Brussels, ULB-VUB, Brussels, Belgium \and 
Machine Learning Group, Universit{\'e} Libre de Bruxelles, Brussels, Belgium \and
Multitel, Mons, Belgium \and
IRIDIA, Universit{\'e} Libre de Bruxelles, Brussels, Belgium \and
ISIA unit, University of Mons, Mons, Belgium
}

\maketitle              
\begin{abstract}
   EXplainable Artificial Intelligence (XAI) aims to help users to grasp the reasoning behind the predictions of an Artificial Intelligence (AI) system. Many XAI approaches have emerged in recent years. Consequently, a subfield related to the evaluation of XAI methods has gained considerable attention, with the aim to determine which methods provide the best explanation using various approaches and criteria. However, the literature lacks a comparison of the evaluation metrics themselves, that one can use to evaluate XAI methods. This work aims to fill this gap by comparing $14$ different metrics when applied to nine state-of-the-art XAI methods and three dummy methods (e.g., random saliency maps) used as references. Experimental results show which of these metrics produces highly correlated results, indicating potential redundancy. We also demonstrate the significant impact of varying the baseline hyperparameter on the evaluation metric values. Finally, we use dummy methods to assess the reliability of metrics in terms of ranking, pointing out their limitations.
   
\end{abstract}













\section{Introduction}

EXplainable Artificial Intelligence (XAI) aims to provide accurate predictions and explanations of these predictions from an AI system in humanly understandable terms~\cite{das2020opportunities}. 
One of the most widely used type of XAI methods in computer vision is attribution-based methods, which provide a weighted set of relevant features concerning the prediction of the model. These methods allow for relevant visualizations through saliency maps and are sometimes referred to as ``saliency methods''. While a broad class of such methods exists, evaluating them remains a significant challenge because of the lack of ground-truth explanations~\cite{samek2017Eva}.
Several properties and requirements (e.g., faithfulness), leading to the formulation of quantitative metrics, have been proposed to assess the quality of saliency methods~\cite{alvarez2018towards,bhatt2020Eva}. 
While the existence of multiple metrics could permit a better overview of the reliability of a given method to be obtained, newly proposed saliency methods are often only compared to existing work using a limited subset of the available evaluation metrics. Furthermore, there is little to no quantitative analysis or detailed investigation into the potential redundancy of XAI evaluation metrics. This paper proposes to fill this gap by studying the concordance and redundancy of XAI evaluation metrics. To do so, we perform statistical analysis on various metrics after evaluating nine state-of-the-art XAI methods for convolutional neural networks (CNN) applied to computer vision and three dummy methods that are not real saliency methods and are used as references for sanity check purposes. In addition, we evaluate the influence of the baseline hyperparameter of the metrics (e.g., black or white pixels to simulate the deletion/insertion of pixels). Previous work has shown that the choice of the baseline has a big influence on the results of XAI methods~\cite{sturmfels2020visualizing}. However, to the best of our knowledge, no work has been done exploring the impact of the baseline for the metrics, nor its impact on the produced scores. 
Finally, thanks to the dummy methods, we discuss the reliability of metrics. The remainder of the paper is organized as follows. Section~\ref{sec:related_work} presents an overview of related work. Section~\ref{sec:experimental_setup} describes the experimental setup that led to the results discussed in Section~\ref{sec:experimental_results}. Section~\ref{sec:conclusion} concludes the paper.

\section{State of the Art}\label{sec:related_work}

\subsection{Saliency Methods}\label{sec:sal_methods}
There is a large variety of saliency methods. This work does not have the ambition to be exhaustive, therefore we will focus on three families only: gradient-based methods, perturbation-based methods, and Class Activation Map (CAM) methods. 
The models considered are CNNs for which the XAI literature is extensive. Saliency maps for new architectures, such as ViT \cite{dosovitskiy2020image} are not entirely interchangeable with respect to CNNs, apart from gradient-based methods and some perturbation methods. This section will evoke the principal characteristics of the methods used in our experiments. For a detailed description of the functioning of the different methods, we refer the reader to the indicated papers specific to each method.

\subsubsection{Gradient-based methods}
Gradient-based methods use the gradient with respect to the inputs (or a modified, backpropagated version thereof), i.e., the derivative of the model output with respect to its input. This derivative produces a saliency map. These methods were introduced in the pioneering work of Simonyan et al.~\cite{Simonyan2014} to highlight the most significant pixels in the decision-making of an image classification neural network. Note that it is the derivative of the preactivations of the softmax. Here, we shall present three representative variants.
\textit{SmoothGrad}~\cite{smilkov2017smoothgrad} was proposed to explicitly reduce noise in attributions. It generates multiple samples by adding Gaussian noise to the original image and averages the calculated gradients to produce a saliency map. \textit{Guided backpropagation}~\cite{springenberg2014striving} is similar to~\cite{Simonyan2014}, with a specific handling of ReLu during the backpropagation. It keeps the non-negative values found in the forward and backward pass when back-propagating through the ReLU functions. \textit{Integrated gradients}~\cite{sundararajan2017axiomatic} linearly interpolates the input image $\bm{x}\in \mathbb{R}^D$ with a baseline image $\bm{x}'$ over an $\alpha \in [0, 1]$ parameter, and averages the gradient for all of these interpolations. The baseline input represents the absence of a feature in the image (e.g., black or uniform, Gaussian noise, blurred image, etc.).

\subsubsection{Perturbation-based Methods}
These methods analyze the evolution of the model's output as a function of variations in the input. They aim to measure whether information lost due to the perturbation is negatively or positively correlated to the prediction score. 
Among the most recent techniques, \textit{RISE}~\cite{petsiuk2018rise} uses random occlusion patterns produced by bilinear upsampling of lower resolution binary masks to perturb the input.
The model produces a probability score for each masked image for each class, the saliency map for the original image being produced for each class of interest by a linear combination of the masks, weighted by the corresponding class scores for each mask.
 
 \subsubsection{CAM Methods}
CAM methods use the activation from a convolutional layer to produce the saliency maps.
The original method~\cite{zhou2016learning} uses the activations of the last convolutional layer and linearly balances them by the weights of a one-layer linear classifier. Since CAM has limited applicability across architectures, \textit{Grad-CAM}~\cite{selvaraju2017grad} was introduced as a generalization, defining the weighting factors for the linear combination of the activations as the average of the gradient flowing through each channel of the last convolutional layer. 
\textit{Score-CAM}~\cite{wang2020score} does not depend on gradients, as it generates the weighting factors to combine the activation map using its forward passing score on the target class. The saliency map is then computed by a linear combination of the obtained scores and activation maps. 
\textit{Poly-CAM}~\cite{englebert2022polycam} recursively multiplexes the high-resolution activation maps in the early layers with upsampled versions of the class-specific activation maps in the last layers. The weighting factors for the multiple CAMs are derived from scores computed similarly to perturbation methods and Score-CAM.
\textit{Layer-CAM}~\cite{jiang2021layercam} gathers CAM from all layers of the CNN by directly multiplying the gradient element-wise with the activation and averaging over the channels instead of performing a linear combination. This allows object localization collection from rough spatial localization to fine-grained details.
\textit{CAMERAS}~\cite{jalwana2021cameras} fuses the activation maps and backpropagated gradients of a layer for different scaled versions of the same input image before performing an element-wise product of these fused activation maps and gradients, similarly to Layer-CAM on the last convolutional layer.

\subsection{Evaluation Metrics}
\label{subsec:evaluation_metrics}
In order to evaluate and compare the different XAI methods presented in Section~\ref{sec:sal_methods}, there are several types of metrics. We consider the four distinct families of such metrics that are applicable to our dataset, based on the property of explainability they aim to characterize, following the formalism proposed in Quantus~\cite{hedstrom2022quantus}. \textit{Faithfulness metrics} measure how explanations follow the predictive behavior of the model. Seven of them are explored in this paper. Faithfulness Correlation~\cite{bhatt2020Eva} partitions the input image into subsets of features whose values are iteratively replaced by a baseline value (e.g., black pixel), and computes the Pearson correlation between the drop in classification probability and the sum of the relevance attribution of a subset. Similarly, Faithfulness Estimation~\cite{alvarez2018towards} calculates the Pearson correlation between the drop in classification probability and the feature relevance. Monotonicity Arya~\cite{arya2019one} measures the extent to which model performance increases (as measured by classification probability) when features of increasing significance are added, whereas Monotonicity Nguyen~\cite{nguyen2020quantitive} follows the same idea, but measures the model performance increase through probability estimation uncertainty. Pixel Flipping~\cite{bach2015pixel}, also called Deletion metric by some authors~\cite{petsiuk2018rise,wang2020score,englebert2022polycam} flips pixels with high relevance scores from the relevance heatmap and looks at the probability score evolution. 
Region Perturbation~\cite{samek2017Eva}) and Selectivity~\cite{montanov2018deep} both extend this methodology to areas of an image. \textit{Robustness metrics} measure the stability of explanations to small input perturbations.
Three of them are used in the following work. The Local Lipschitz Estimate~\cite{alvarez2018robustness} measures the consistency in explanations for adjacent samples. Max-Sensitivity and Avg-Sensitivity~\cite{yeh2019fidelity} quantify how explanations change when inputs are infinitesimally perturbed, through a Monte Carlo sampling-based approximation. \textit{Complexity metrics} measure explanation conciseness. Three of them are compared in the experiments. Sparseness~\cite{chalasani2020concise} uses the Gini index to measure whether only highly-attributed features are predictive of the model output. Complexity~\cite{bhatt2020Eva} measures the entropy of the fractional contributions to the total attribution, whereas Effective Complexity~\cite{nguyen2020quantitive} looks at how many attributions exceed a certain threshold. \textit{Randomization metrics} measure model deterioration as a function of parameter randomization. Two methods are explored. Model Parameter Randomization~\cite{abedayo2018sanity} quantifies the similarity between original explanations and explanations from sequential randomization of successive model layers. The Random Logit Test~\cite{sixt2020explanations} computes the distance between the original explanation and the explanation obtained for another random class.

\subsection{Benchmarking Saliency Evaluation  Metrics}\label{sec:metric_benchmarking}

Previous works that quantitatively evaluate metrics of saliency methods are mainly found in XAI literature surveys.
For example, general overviews of XAI method evaluations are available~\cite{burkart2021survey}. Beyond descriptions, Zhou et al.~\cite{zhou2021evaluating} couple XAI methods with corresponding evaluation metrics.
In this paper, we assume families of metrics as described in Section~\ref{subsec:evaluation_metrics}. While several works in the literature discuss XAI evaluation metrics, very little is known about potential redundancies between metrics in the same family or not. Noteworthy is the work from Li et al.~\cite{Li2021}, which presents a review that encompasses a quantitative evaluation of seven saliency methods. These methods are compared using three types of metrics: faithfulness, localization and robustness. Li et al.~\cite{Li2021} point out that RISE and Grad-CAM perform well for most metrics. However, they also indicate that no particular method is consistently ranked as best with all metrics. Our work goes beyond the abovementioned approaches by offering an inquiry focused on metrics rather than methods. We evaluate and identify representative and potentially redundant metrics to obtain a comprehensive view of a method in terms of XAI evaluation. 

\section{Experimental Setup to Compare Evaluation Metrics}
\label{sec:experimental_setup}

This section describes the experimental setup to obtain the results presented in Section~\ref{sec:experimental_results}. Section~\ref{subsec:data_model} presents the image data set and the deep learning models used to obtain the image classifications to be explained. Sections~\ref{subsec:methods} and~\ref{subsec:metrics} introduce the selected saliency methods and evaluation metrics, respectively. Finally, Sections~\ref{subsec:processing_results} and~\ref{subsec:analyzing_methods} describe how we process and analyze the results. 
The code is made available for reproducibility\footnote{
\url{https://github.com/multitel-ai/Evaluation-XAI-Metrics-GD6-TRAIL}.
}.

\subsection{Data Set and Model for Image Classification}\label{subsec:data_model}
When evaluating saliency methods, it is common to use the ImageNet (2012 ILSVRC) validation set~\cite{ILSVRC15} (e.g.,\cite{wang2020score,petsiuk2018rise}) or the CIFAR10 dataset~\cite{krizhevsky2009learning} (e.g., \cite{chalasani2020concise,yeh2019fidelity,alvarez2018towards}).
In accordance with~\cite{wang2020score}, we use a subset of $2000$ randomly selected images from ImageNet, and we employed two widely-known pretrained models (ResNet-50~\cite{he2016deep} and VGG-16~\cite{Simonyan15}) from the PyTorch model zoo to obtain predictions, which are explained thanks to the methods in Section~\ref{subsec:methods}. For CIFAR10, the 2000 first images of the test dataset were utilized with a ResNet50 model pretrained on CIFAR10 available at~\cite{huy_phan_2021_4431043}.

\subsection{Selected XAI Methods for Image Classification Explanation}\label{subsec:methods}
For image classification explanation, we use state-of-the-art XAI methods that include Integrated Gradient~\cite{sundararajan2017axiomatic}, SmoothGrad~\cite{smilkov2017smoothgrad}, Guided Backpropagation~\cite{springenberg2014striving}, Grad-CAM~\cite{selvaraju2017grad}, Score-CAM~\cite{wang2020score}, Layer-CAM~\cite{jiang2021layercam}, Poly-CAM~\cite{englebert2022polycam}, RISE~\cite{petsiuk2018rise} and {CAMERAS}~\cite{jalwana2021cameras}.
Three dummy saliency maps were added to the methods used to compare metrics: a randomly generated map (sampling a standard uniform distribution $U(0,1)$ for each pixel), a Sobel filter
and a map produced from a two-dimensional centered Gaussian ($\bm{\mu}=0$, $\Sigma=I$). These three methods can be seen as an ``ablation study'' on metrics, which is a new contribution of our study since it has not been investigated before to the best of our knowledge. Indeed, they allow the comparison of rankings produced for a metric according to the expected ranking of dummy methods (w.r.t. the evaluated property such as faithfulness). This is shown in Section~\ref{subsec:sanity_check}, on the reliability of the metrics.





\subsection{Selected Evaluation Metrics to Compare XAI Methods}\label{subsec:metrics}
In order to evaluate the above XAI methods, evaluation metrics were selected from those introduced in Section~\ref{subsec:evaluation_metrics}. For faithfulness metrics, these include Faithfulness Correlation, Faithfulness Estimate, Pixel Flipping, Selectivity, Monotonicity Nguyen and Monotonicity Arya. We did not consider the Region Perturbation metric~\cite{samek2017Eva}, as it behaves in the same way as the Selectivity metric~\cite{montanov2018deep}.
For robustness metrics, the selected metrics include the Local Lipschitz Estimate, Max-Sensitivity and Avg-Sensitivity, that were applied with ten Monte Carlo samples (to have a reduced computational burden\footnote{\label{footnote2}For example, evaluating Poly-CAM on Avg-Sensitivity with $10$ MC samples took approximately 22 hours with five parallel jobs on an NVIDIA Geforce GTX 1080 Ti.}). From the group of complexity metrics, those selected were 
Sparseness, Complexity and Effective Complexity. Random Logit and Model Parameter Randomization were included as randomization metrics. The latter was used with a top-to-down randomization scheme~\cite{abedayo2018sanity}. We used the metric implementations from Quantus~\cite{hedstrom2022quantus}.
Finally, one important consideration when evaluating faithfulness metrics is the simulation of \textit{feature removal}, which is technically done by replacing a pixel value to a \textit{baseline} value (e.g., black, white or random). We studied the influence of this baseline hyperparameter on the scores generated by faithfulness metrics. As explained in Section~\ref{subsec:evaluation_metrics}, faithfulness metrics measure how explanations relate to the predictive behavior of the model by perturbing parts of each input with a chosen value. In our experiments, we used four baselines: black, white, random and uniform values as implemented in Quantus.

\subsection{From Method Evaluation Scores to a Matrix of Scores}\label{subsec:processing_results}
As a result of running all the aforementioned methods and evaluating them with the 14 listed metrics, we obtained, for each method either a score or a list of scores for each image, depending on the metric. From the listed metrics, only Pixel Flipping, Selectivity and Model Parameter Randomization returned a list of scores for each image. In these cases, it is common to aggregate the result by computing the AUC of the list of scores~\cite{wang2020score}.
As our goal is to understand how metrics behave in relation to each other, we want to obtain a matrix $\bm{X}$ whose rows and columns correspond to methods and metrics, respectively. Each element of this matrix $X_{ij}$ is obtained by averaging the scores of the $j$-th metric applied to the outputs (i.e., 2000 images) of the $i$-th method, similar to what Li et al.~\cite{Li2021} did.

\subsection{Comparison of Evaluation Metrics through Method Scores}\label{subsec:analyzing_methods}

In order to investigate pairwise comparisons of metrics, we compute the correlation coefficients on the scores produced by these metrics (i.e., the columns $X_{.j}$ of $\bm{X}$ described above). The objective is to identify potential pairs of redundant metrics. Since we have potentially non-linear relationship between scores, an adequate solution to assess correlations is Kendall's \(\tau_b\) rank coefficients~\cite{kendallstau}.
The use of \(p\)-values with Kendall's \(\tau_b\) can reveal significant correlations.
In the rest of this paper, Kendall's \(\tau_b\) will simply be referred to as Kendall's \(\tau\). By multiplying pairwise comparisons, the risk of wrongly rejecting a null hypothesis (i.e., the absence of correlation) increases. To prevent this, we used Holm-Bonferroni-corrected  \(p\)-values~\cite{holm1979simple} with a family-wise error rate of $0.05$. Beyond significance, it is important to remain cautious about the interpretation of correlations in the context of XAI methods. Correlated scores can still contain some major differences. Only (near) perfectly correlated metrics, i.e., with a Kendall \(\tau\) superior to \(0.9\), will be considered as potentially redundant. Based on the definition of Kendall's \(\tau\)
\begin{equation}
    \tau = \frac{(p - q)}{\sqrt{(p + q + t)(p + q + u)}},
\end{equation}
with $p$ the number of concordant pairs, $q$ the number of discordant pairs, $t$ the number of ties in the first ranking, and $u$ the number of ties in the second ranking, and on the hypothesis that there are no ties in any of the two compared rankings ($t = u = 0$), the percentage of concordant pairs can be derived from Kendall's $\tau$. The threshold of $0.9$ that we use corresponds to $95\%$ matching pairs (i.e., fewer than 3 discordant pairs).

\section{Experimental Results for Evaluation Metrics}\label{sec:experimental_results}

This section analyzes the correlations of metrics, the influence of hyperparameters of metrics and the reliability of evaluation metrics.

\subsection{On the Correlation of Evaluation Metrics}\label{subsec:compementary_metrics}
\begin{figure*}[!htbp]
    \begin{subfigure}{0.53\textwidth}
    \includegraphics[width=\linewidth]{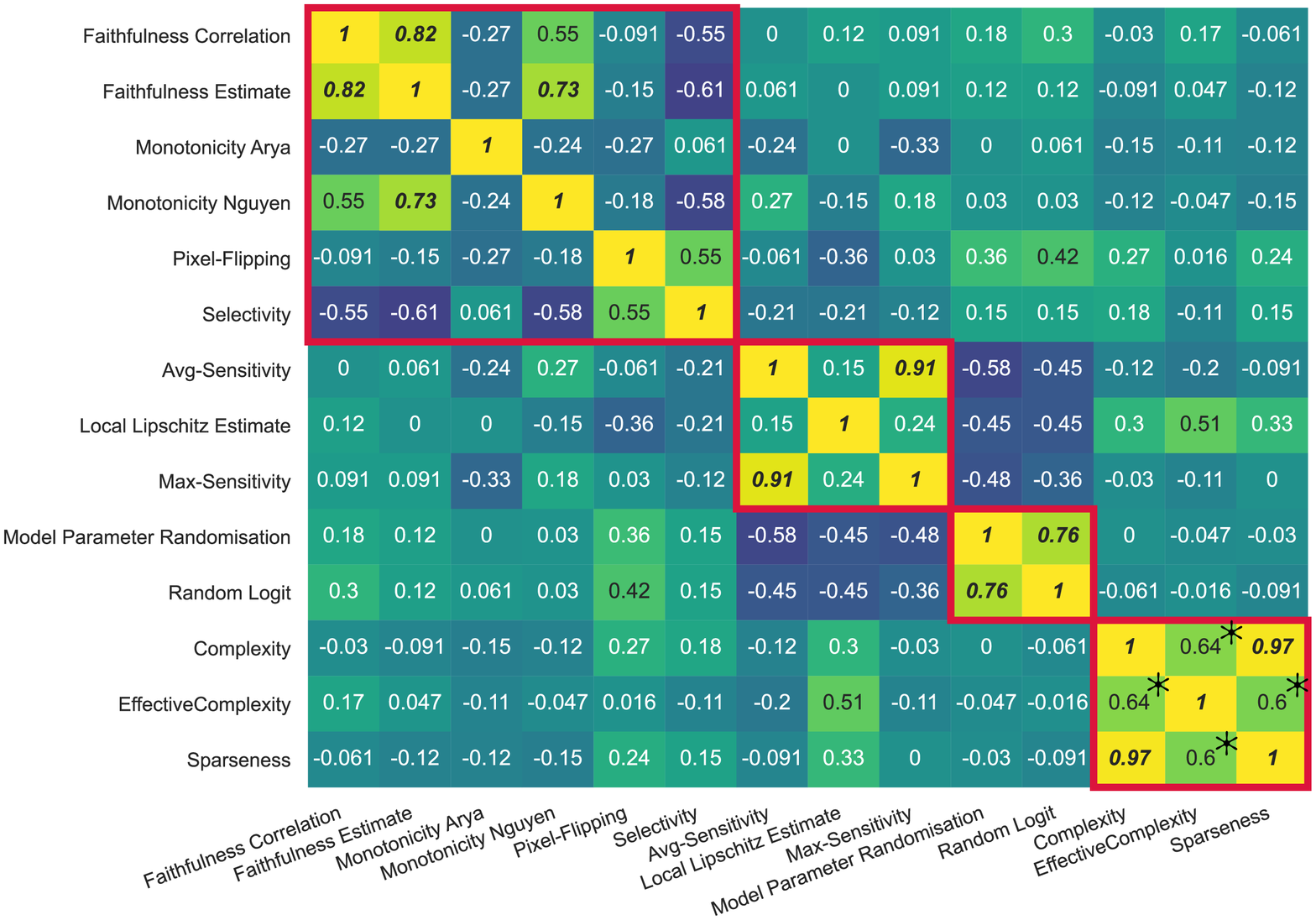}
    \caption{ResNet-50 on ImageNet}
    \label{fig:corr_matrix_all_metrics_resnet50}
    \end{subfigure}
    \hspace{-0.2cm}
    \begin{subfigure}{0.469\textwidth}
    \includegraphics[width=\linewidth]{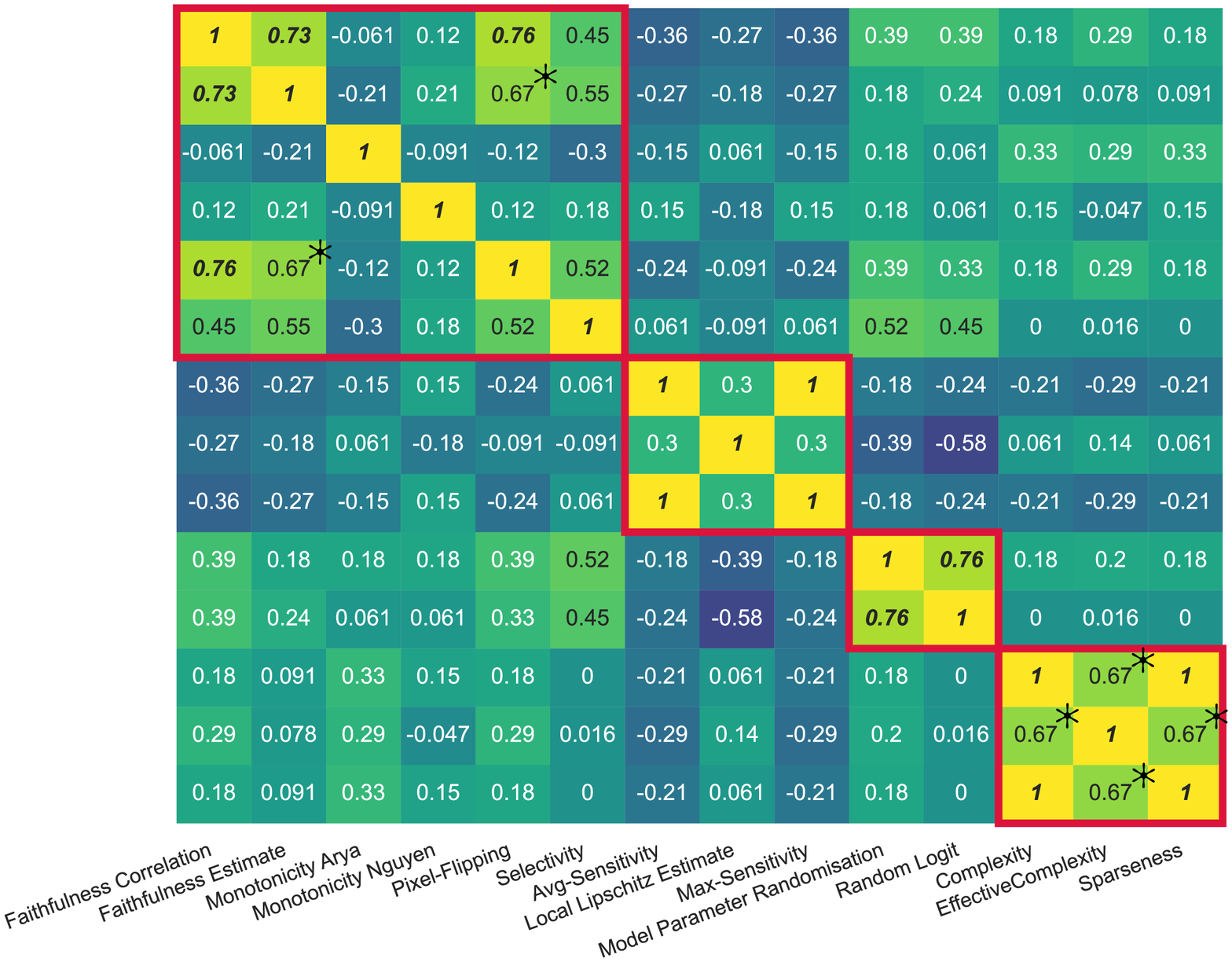}
    \caption{VGG-16 on ImageNet}
    \label{fig:corr_matrix_all_metrics_vgg16}
    \end{subfigure}
    \caption{Kendall's \(\tau\) correlation coefficients for all metrics, with black as default baseline for faithfulness metrics. Groups of metrics are highlighted in red with from left to right Faithfulness, Randomisation, Robustness and Complexity~\protect\footnotemark[3]. Intra-group significant values are stared (``*''), based on Holm-Bonferroni-corrected \(p\)-values by taking into account the corresponding group sub-matrix.} 
    \label{fig:corr_matrix_all_metrics_}
\end{figure*}

\begin{figure}
\begin{minipage}[!htbp]{0.511\textwidth}
    \includegraphics[width=\linewidth]{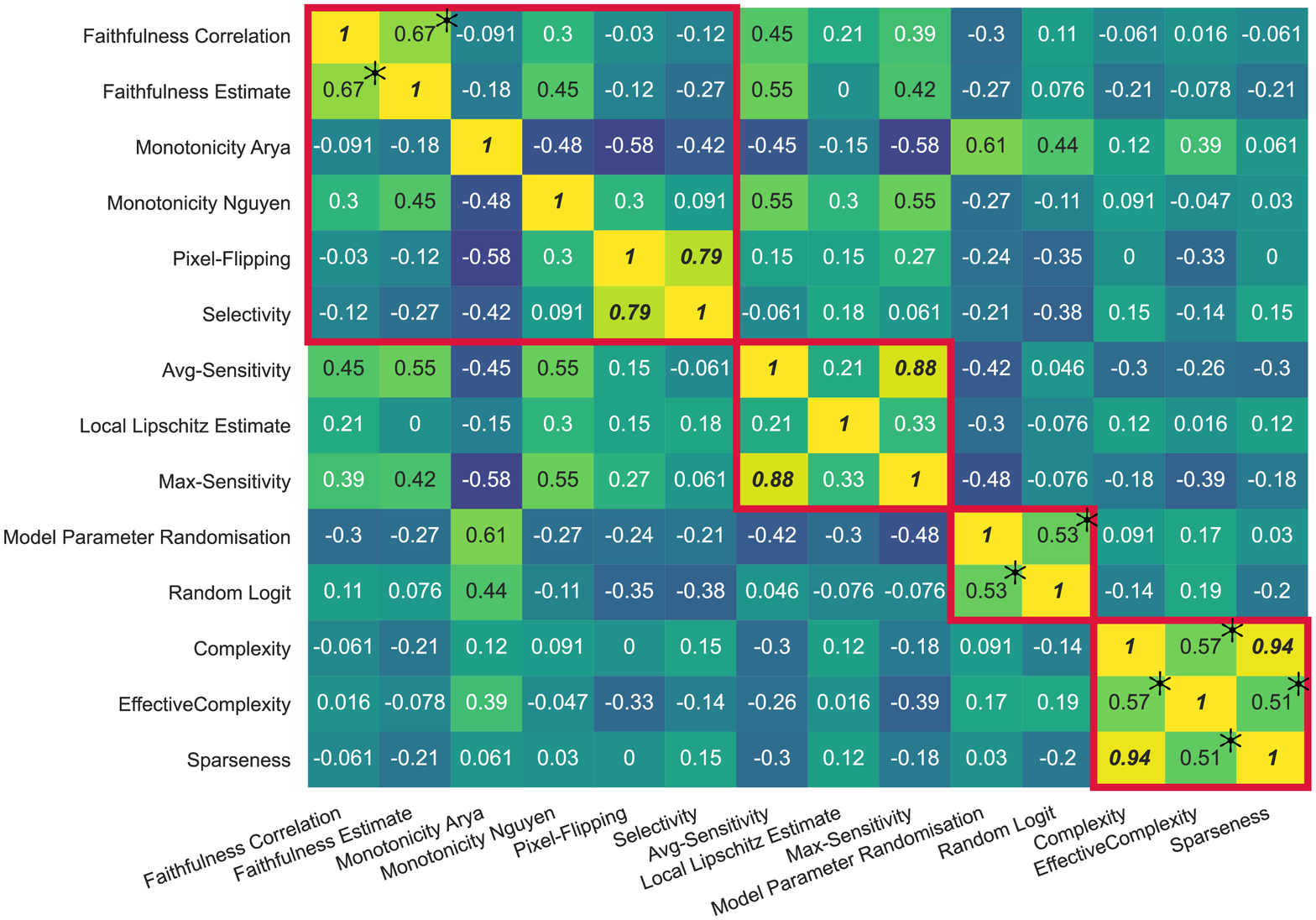}
    \caption{Kendall's \(\tau\) correlation coefficients for all metrics using ResNet-50 on CIFAR10, with black as default baseline for faithfulness metrics\protect\footnotemark[3]. Intra-group significant values are stared (``*''), based on Holm-Bonferroni-corrected \(p\)-values by taking into account the corresponding group sub-matrix.
    }
    \label{fig:corr_matrix_all_metrics_cifar}
\end{minipage}
 \hfill
\begin{minipage}[!htbp]{0.458\textwidth}
    \begin{subfigure}{0.48\textwidth}
        \includegraphics[width=\columnwidth]{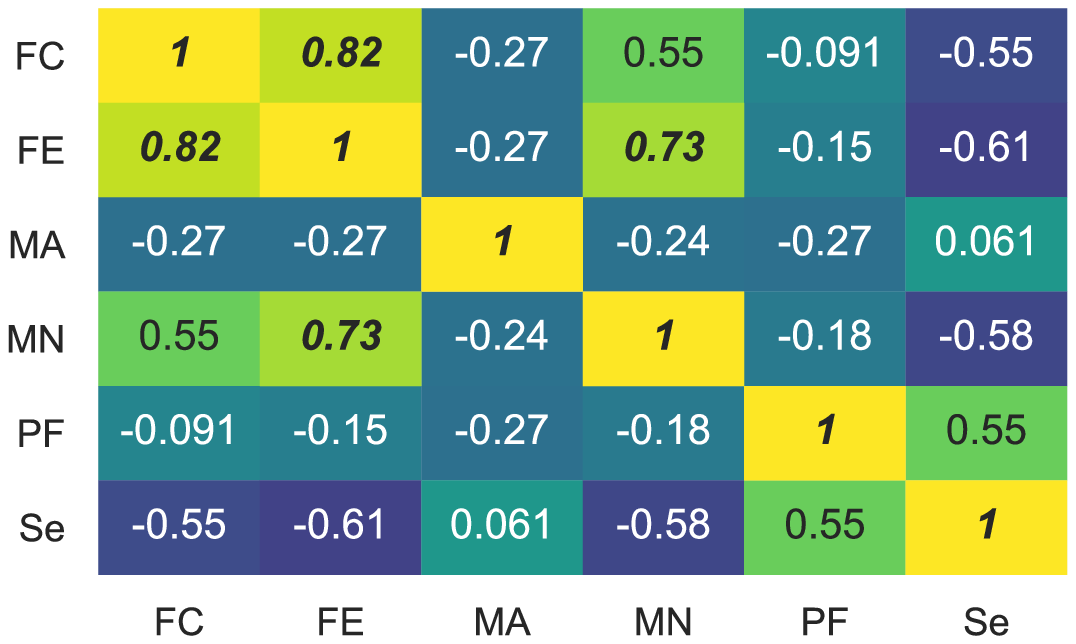}
        \caption{Black}
        \label{corr_matrix_faithfulness_black}
    \end{subfigure}
    \hspace{0.01cm}
    \begin{subfigure}{0.49\textwidth}
        \includegraphics[width=\columnwidth]{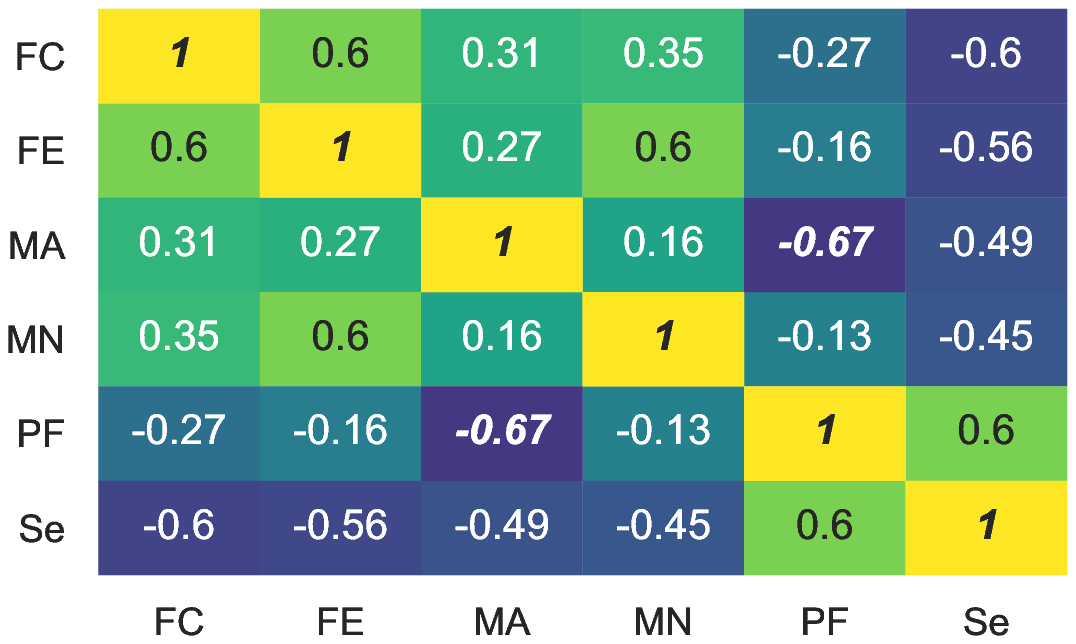}
        \caption{White}
        \label{corr_matrix_faithfulness_white}
    \end{subfigure}
    \hspace{0.1cm}
    \begin{subfigure}{0.475\textwidth}
        \includegraphics[width=\columnwidth]{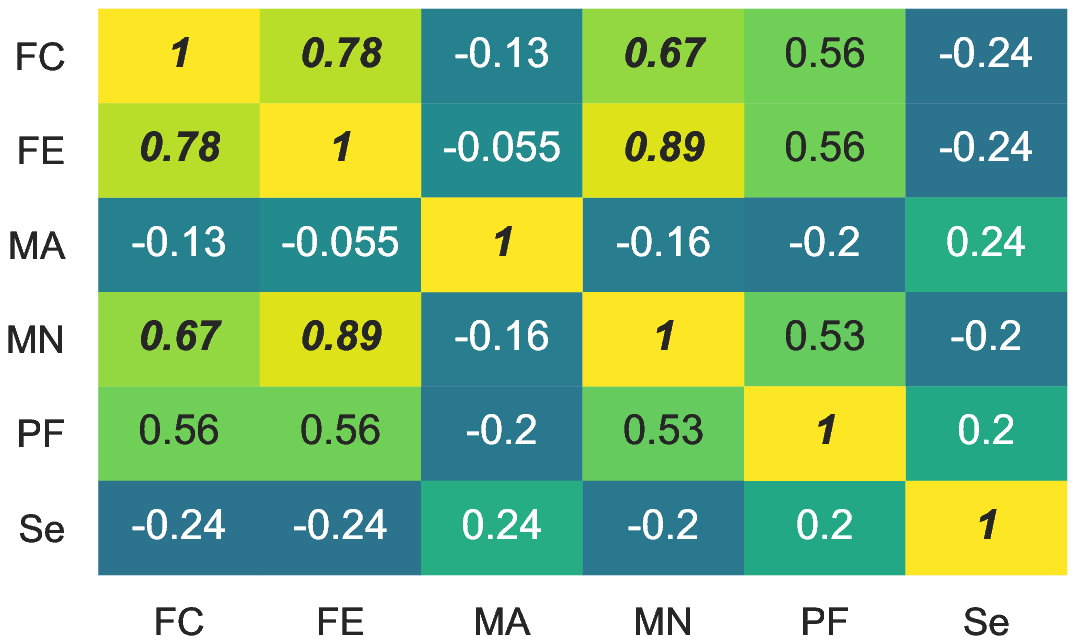}
        \caption{Random}
        \label{corr_matrix_faithfulness_random}
    \end{subfigure}
    \hspace{0.1cm}
    \begin{subfigure}{0.48\textwidth}
        \includegraphics[width=\columnwidth]{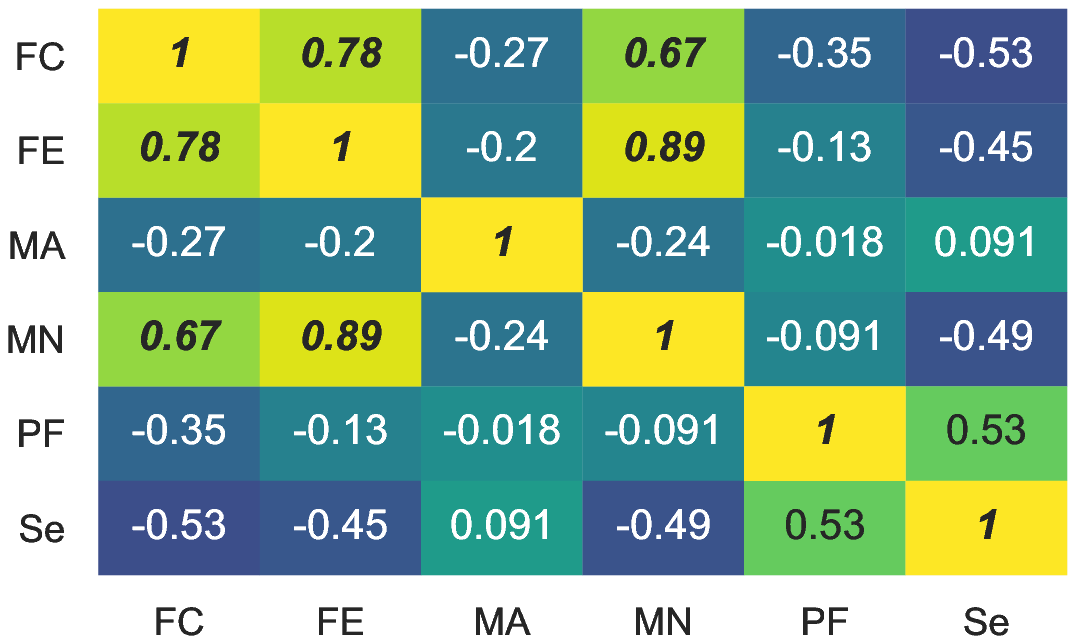}
        \caption{Uniform}
        \label{corr_matrix_faithfulness_uniform}
    \end{subfigure}
    \caption{Kendall's \(\tau\) correlations for faithfulness metrics using different baselines on ImageNet\protect\footnotemark[3] using Resnet-50: Faithfulness Correlation (FC), Faithfulness Estimate (FE), Monotonicity Arya (MA), Monotonicity Nguyen (MN), Pixel Flipping (PF) and Selectivity (Se).}
    \label{fig:corr_matrices_faithfulness_same_baseline_diff_metrics}
\end{minipage}
    
\end{figure}

Given pairs of metrics of the same type, we investigate which of these pairs achieves possible agreement. First, we find two couples of metrics that can be considered potentially redundant (based on the threshold of $0.9$ defined in Section~\ref{subsec:analyzing_methods}).
For \textit{complexity metrics} (bottom right red square in Fig.~\ref{fig:corr_matrix_all_metrics_resnet50},~\ref{fig:corr_matrix_all_metrics_vgg16} and~\ref{fig:corr_matrix_all_metrics_cifar}), the Sparseness and Complexity metrics are significantly correlated with a Kendall's \(\tau\) correlation coefficient of $0.97$ (ResNet-50) and $1.0$ (VGG-16) on ImageNet, and $0.94$ (ResNet-50) on CIFAR10. 
For \textit{robustness metrics} (second red square in Fig.~\ref{fig:corr_matrix_all_metrics_resnet50}, ~\ref{fig:corr_matrix_all_metrics_vgg16} and ~\ref{fig:corr_matrix_all_metrics_cifar}), the results show that Max-Sensitivity and Avg-Sensitivity are correlated for CIFAR10 on ResNet-50 (($\tau$=$0.88$) and largely correlated for both models on ImageNet ($\tau$=$0.91$ for ResNet-50 and $\tau$=$1.0$ for VGG-16) where they can be considered potentially redundant. Indeed, it is intuitive that Avg-Sensitivity and Max-Sensitivity are potentially redundant because the more similar the perturbed values in the neighborhood of the input are, the more the average values will be correlated to the maximum value. The choice is left to users if they want to give importance to some maximum perturbation outliers and prefer the Avg-Sensitivity rather than the Max-sensitivity. 

After analyzing the correlations above $0.9$, we have the metrics that are significantly correlated below the $0.9$ threshold defined in Section~\ref{subsec:analyzing_methods}.
For \textit{randomization metrics} (third red square in Fig.~\ref{fig:corr_matrix_all_metrics_resnet50},~\ref{fig:corr_matrix_all_metrics_vgg16} and~\ref{fig:corr_matrix_all_metrics_cifar}), the results show that there is a moderately significant intra-group correlation between Model Parameter Randomization and Random Logit, i.e., $0.76$ (ResNet-50) and $0.76$ (VGG-16) on ImageNet, and $0.53$ (ResNet-50) on CIFAR10. This suggests that even though the functioning of the randomization metric is significantly different (layer randomization versus explanation class randomization), explainability methods being sensitive to both types of randomization may be close.


Finally, for \textit{faithfulness metrics} with black as a default baseline, there are no pairs of metrics that are significantly correlated on all models and datasets. We can only observe that in the case of ImageNet on both the Resnet-50 and VGG-16 models (Fig.~\ref{fig:corr_matrix_all_metrics_}) as well as on several baselines (Fig.~\ref{fig:corr_matrices_faithfulness_same_baseline_diff_metrics}), Faithfulness Correlation and Faithfulness Estimate are significantly correlated ($\tau$=$0.82$ for Resnet-50, $\tau$=$0.73$ for VGG-16). Furthermore, Pixel Flipping and Selectivity are significantly correlated on CIFAR10 ($\tau$=$0.79$ on Resnet-50). We suggest that users use these metrics to analyze methods having similar maps (e.g., Gradient-based methods only or CAM methods only), because methods such as the Gradient-based methods produce sparser saliency maps (with high-frequency pixels) that will, most of the time, outperform any other methods.
Beyond an intra-group analysis, it can be seen that there is no significant inter-group correlation between metrics for both models. This result is consistent with the fact that different types of metrics evaluate different properties of saliency methods.

\begin{figure}[!htbp]
    \begin{subfigure}{0.16\linewidth}
        \includegraphics[width=\columnwidth]{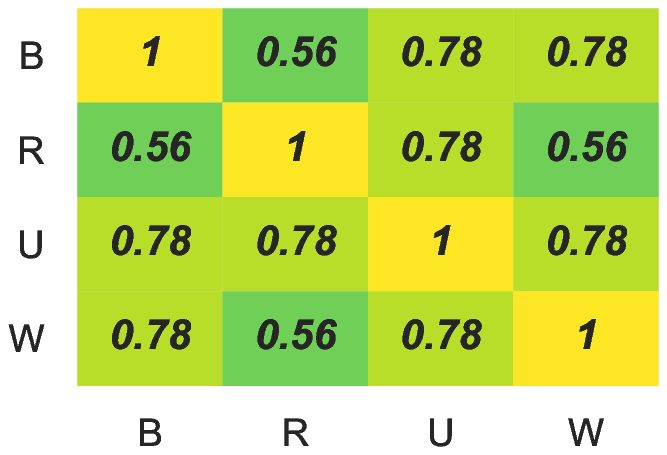}
        \caption{Faith. Corr.}
        \label{corr_matrix_faithfulness_correlation}
    \end{subfigure}
    \hspace{-0.1cm}
    \begin{subfigure}{0.16\linewidth}
        \includegraphics[width=\columnwidth]{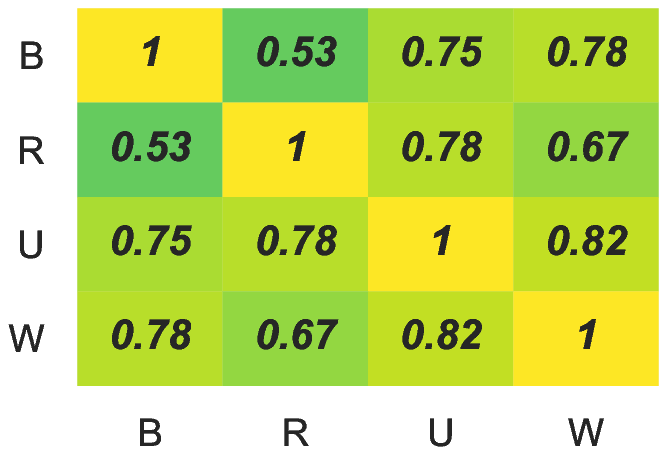}
        \caption{Faith. Est.}
        \label{corr_matrix_faithfulness_estimate}
    \end{subfigure}
    \hspace{-0.1cm}
    \begin{subfigure}{0.16\linewidth}
        \includegraphics[width=\columnwidth]{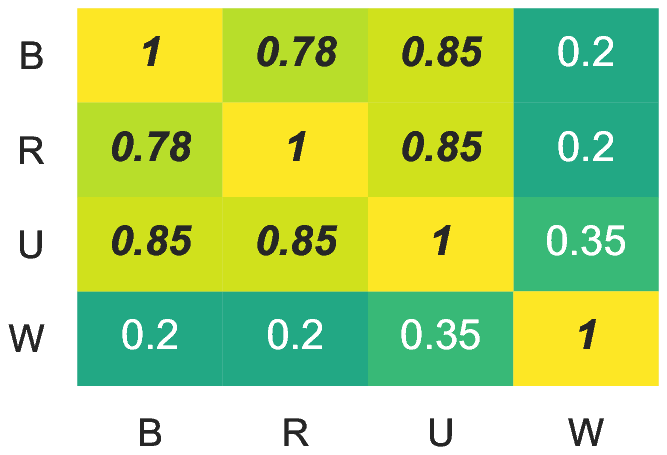}
        \caption{M. Arya}
        \label{corr_matrix_monotonicity_arya}
    \end{subfigure}
    \begin{subfigure}{0.16\linewidth}
        \includegraphics[width=\columnwidth]{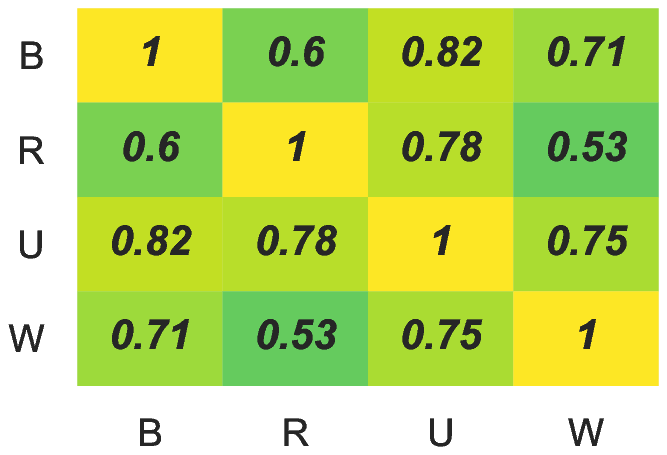}
        \caption{M. Nguyen}
        \label{corr_matrix_monotonicity_nguyen}
    \end{subfigure}
    \begin{subfigure}{0.16\linewidth}
        \includegraphics[width=\columnwidth]{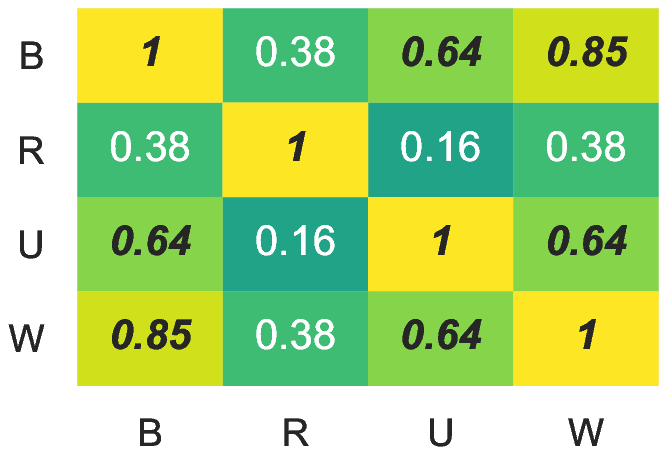}
        \caption{Pixel Flipp.}
        \label{corr_matrix_pixel_flipping}
    \end{subfigure}
    \begin{subfigure}{0.16\linewidth}
        \includegraphics[width=\columnwidth]{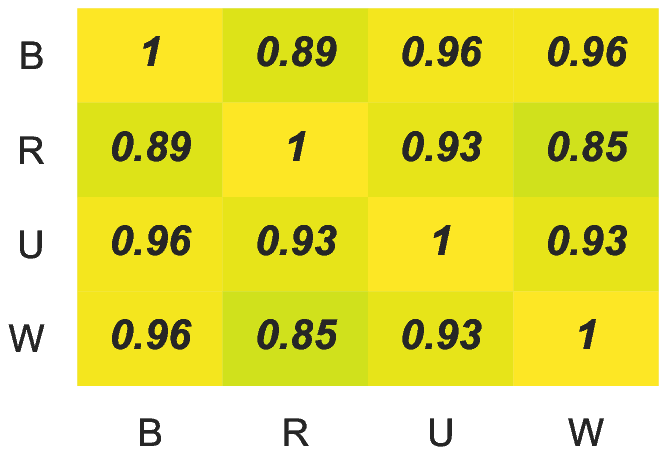}
        \caption{Selectivity}
        \label{corr_matrix_selectivity}
    \end{subfigure}
    \caption{Kendall \(\tau\) correlation coefficients for the faithfulness metrics and different baselines on ImageNet~\protect\footnotemark[3]
    . Black (B), Random (R), Uniform (U), White (W) on ResNet-50.}
    \label{fig:corr_matrices_faithfulness_same_metric_diff_baselines}
\end{figure}



\subsection{On the Hyperparameters of Evaluation Metrics} 
In this section, we evaluate and discuss correlations for faithfulness metrics (with the ResNet-50 model) according to baseline values. All previous results were reported using black as the default baseline (value given to a pixel to simulate the \textit{feature removal}) for all the faithfulness metrics that relied on this hyperparameter. We explore the impact of the baseline for the metrics and its impact on the produced scores.

Fig.~\ref{fig:corr_matrices_faithfulness_same_baseline_diff_metrics} shows Kendall's $\tau$ correlation coefficients for the faithfulness metrics on different baselines. According to this figure, there are significant variations in the correlations calculated according to the baselines, such as the correlation between Monotonicity Arya and Pixel Flipping for a white and uniform baseline (-0.67 and -0.01 respectively). 
This is further emphasized in the results of Fig.~\ref{fig:corr_matrices_faithfulness_same_metric_diff_baselines}, illustrating Kendall's \(\tau\) correlation coefficients for each baseline and each faithfulness metric. Apart from Selectivity which shows resilience with correlation coefficients consistently over $0.85$, the faithfulness metrics demonstrate high variability in the concordance of the scores depending on the baseline: all correlations are below $0.9$, implying at least 3 discordant pairs in the scores according to the baseline (sometimes even below $0.2$ as for Pixel Flipping or Monotonicity Arya). Therefore, we recommend that new authors use multiple baselines in experiments using faithfulness metrics for a fair comparison. Especially since faithfulness metrics are, to our knowledge, the most widely used in the state of the art to evaluate XAI methods. Otherwise, it would mean drawing conclusions about potentially better XAI methods for a problem or dataset that may prove unreliable, as shown by the variability of results for different baselines.


\interfootnotelinepenalty=10000\footnotetext[3]{Significant values are in italic bold, based on Holm-Bonferroni-corrected \(p\)-values by taking into account the complete matrix.}
\subsection{On the Reliability of Metrics in Terms of Rankings}\label{subsec:sanity_check} 
\begin{figure}[!htbp]
  \centering
    \includegraphics[width=\textwidth]{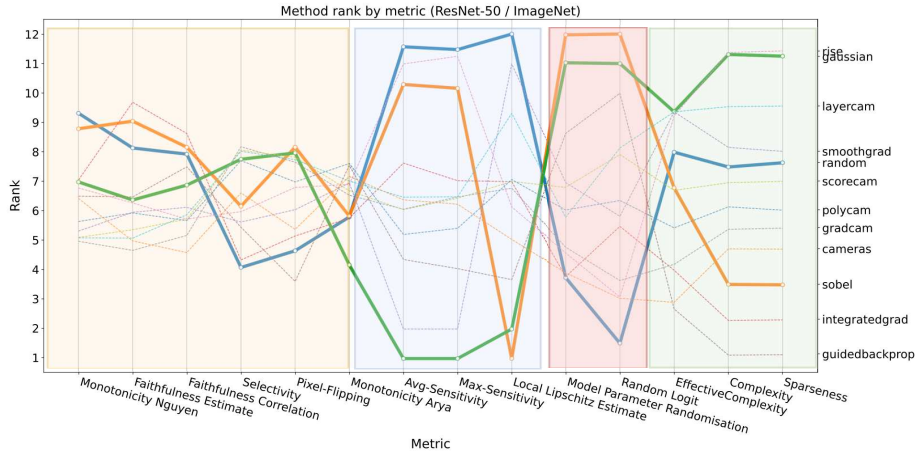}
    \caption{Ranking of XAI methods with respect to metrics for ResNet-50 model on ImageNet (lower rank is better).}
    \label{fig:rank_bump_resnet50}
\end{figure}

\begin{figure*}[!htbp]
    \centering
    \includegraphics[width=1\textwidth]{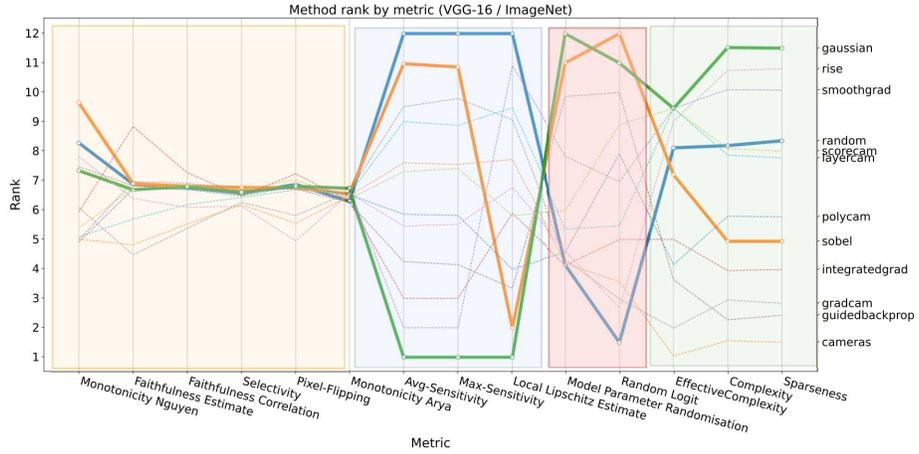}
    \caption{Ranking of XAI methods with respect to metrics for VGG16 model on ImageNet (lower rank is better).}
    \label{fig:rank_bump_vgg16}
\end{figure*}

\begin{figure*}[!htbp]
    \centering
    \includegraphics[width=1\textwidth]{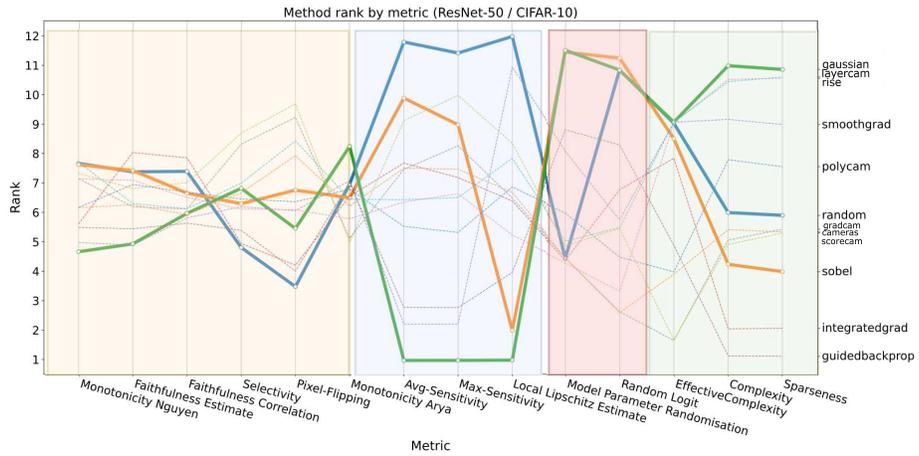}
    \caption{Ranking of XAI methods with respect to metrics for Resnet-50 model on CIFAR10 (lower rank is better).}
    \label{fig:rank_bump_resnet_cifar10}
\end{figure*}

This section explores how the ranking attributed by evaluation metrics to the three dummy explanation methods defined in Section~\ref{subsec:methods} (Sobel, Gaussian and Random) compares with the ranking of state-of-the-art saliency methods. As the former are not valid explainability methods, their ranking in comparison to the ranking of the latter can unveil the strengths and weaknesses of the evaluation metrics. As each family of evaluation metrics is different, some may give dummy methods a low ranking, while others may provide a high ranking. This is discussed in detail below based on Fig.~\ref{fig:rank_bump_resnet50}, Fig.~\ref{fig:rank_bump_vgg16}, and Fig.~\ref{fig:rank_bump_resnet_cifar10}, which show the average ranking (rank $1$ being the best) of the methods for each metric (based on the ranking for each image given by the metric scores) according to the model and dataset. We start the analysis by families of metrics from right to left in the figures.

For \textit{complexity} metrics, intuitively the complexity of the Sobel method depends on the input image, while that of Random is variable by definition. Thus, the ranking of these two dummy methods is not defined. However, the Gaussian method, with its constant and large map, should and does achieve a consistently poor ranking. The results obtained are consistent for all models and datasets.

For \textit{randomization} metrics, Model Parameter Randomization (MPR) and Random Logit (RL) give a bad score to Sobel and Gaussian methods. This is the expected result since they do not depend on the model weights (MPR metric) or on the explanation (RL metric), and thus do not change with the randomization of the model or of the explanation class. On the other hand, as the Random dummy method produces a different result according to the model randomization (MPR metric) or the explanation class randomization (RL metric), it achieves an artificially good ranking by mimicking the big influence of the randomization. Note that it also shows that these metrics can be fooled by a nondeterministic explainability method. One result is, however, inconsistent between datasets because the Random Logit metric gives a very good ranking to the Random method with CIFAR10 (Fig.~\ref{fig:rank_bump_resnet_cifar10}).

For \textit{robustness} metrics, we expect Sobel and Random to have poor rankings, because on the one hand, the maps are completely different each time for Random which makes them very unstable, and on the other hand Sobel detects new edges, being sensitive to local perturbations. Since Gaussian is insensitive to any perturbations of the input, it provides trivial constant explanations and is the optimal choice as stated in the authors' paper~\cite{yeh2019fidelity}. This is what turns out to be the case for Max-Sensitivity and Avg-Sensitivity. However, the Local Lipschitz Estimate rates Sobel as one of the best methods, which is not coherent with the purpose of the metric.

For \textit{faithfulness} metrics, the general goal is that perturbing/adding the most significant pixels or areas of an image has the most considerable effect on the prediction for the best explainability method (and vice versa). However, in addition to being the group of metrics with the least strong correlations in Section~\ref{subsec:compementary_metrics}, we observe that across the three figures, it is also the only family where the results do not agree at all for different models and datasets. All XAI methods are extremely close in rankings for VGG-16, and we observe a lot of crossovers between rankings for all metrics in the other two figures. In terms of dummy methods, since none of the three dummy methods aim to select the most significant pixels, we initially expect them to be consistently ranked poorly. However, we find that Pixel-Flipping and Selectivity assign a good ranking to the random method using Resnet50, or that Gaussian assigns the best or worst ranking for Monotonicity Arya depending on the dataset used. Overall, there is no faithfulness metric that consistently gives a bad ranking to the three dummy methods for all models and datasets used, contrary to what one would logically expect from a reliable method.

\section{Discussion and Conclusion}\label{sec:conclusion}

Despite the existence of multiple quantitative metrics assessing the quality of saliency methods~\cite{alvarez2018towards,bhatt2020Eva}, the evaluation of these methods remains challenging in the absence of ground-truth explanations~\cite{samek2017Eva}. There are also very few objective criteria for choosing methods and metrics. To the best of our knowledge, no previous work exists that focuses on metric comparison instead of saliency method comparison. This paper compares $14$ metrics by evaluating nine state-of-the-art XAI methods, applied to $2000$ images from the ImageNet validation set and $2000$ images from the CIFAR10 test set, using pretrained CNN models.

We studied the pairwise correlation of metrics through Kendall's \(\tau_b\), followed by an analysis of the baseline hyperparameter for the faithfulness metrics, and finally a reliability analysis of the ranking given by the metrics to the XAI methods with the addition of 3 dummy methods used to detect potential issues. From experimental results, we unveil for complexity metrics that Sparseness and Complexity are very highly correlated on all datasets and therefore potentially redundant with each other to represent complexity. For robustness metrics, we find out that Max- and Avg-Sensitivity are strongly correlated. Moreover, as Local Lipschitz Estimate rated unexpectedly highly the Sobel dummy method in the reliability analysis, it would favor the choice of either Max- or Avg-Sensitivity to represent robustness metrics. Model Parameter Randomization and Random Logit demonstrate a significant correlation between each other without being redundant in the randomization metrics, with no consistent flaws found by the use of dummy methods, both representing well different aspects of the randomization. Last, though faithfulness metrics are arguably the most widely used group in the state-of-the-art, this metric group does not find any redundant metrics or significantly correlated ones on all datasets. The reliability analysis shows that the ranks given to the methods vary enormously across datasets and models, and that for these we also cannot find a faithfulness method that gives the assumed rank (worst rank) to the dummy methods used. All this is in addition to the fact that the ranks given by the faithfulness metrics vary significantly depending on the baselines used as shown in our experiments. To draw strong conclusions in experimental analysis, it is essential to include a variation of this hyperparameter to avoid wrong conclusions.

Since there is no ground truth for explainability, the families of metrics each evaluate a different aspect of an explainability method. In real experiments, between the families, it is up to the user to decide which types of metrics they want to focus on (e.g., favor robustness over complexity). Within a family, we show which metrics are correlated and may have less interest to compute in common. In addition, the flaws identified using dummy methods indicate to the user which metrics they may want to avoid.
Future work could extend the paper to analyze reliability under vision transformers or explore the impact of other hyperparameters than the baseline (e.g., the size of patches).

\section{Acknowledgments}
The authors acknowledge the various sources of funding for this research. Julien Albert, Nassim Versbraegen, and Miriam Doh are supported by the Service Public de Wallonie Recherche under grant n°2010235-ARIAC by DIGITALWALLONIA4AI. Nassim Versbraegen is also funded by Innoviris Joint R\&D project Genome4Brussels [2020 RDIR 55b to N.V.]. The Research Foundation for Industry and Agriculture, National Scientific Research Foundation (FRIA-FNRS) funded this research as a grant attributed to Alexandre Englebert, consisting in Ph.D. financing. Geraldin Nanfack is funded by the EOS-VeriLearn project number 30992574 of the Fonds de la Recherche Scientifique (F.R.S-FNRS). Sédrick Stassin thanks the support of the E-origin project funded by the Walloon Region within the pole of logistics in Wallonia.\\

Computational resources have been provided by the supercomputing facilities of the Université catholique de Louvain (CISM/UCL) and the Consortium des Equipements de Calcul Intensif en Fédération Wallonie Bruxelles (CECI) funded by the Fond de la Recherche Scientifique de Belgique (F.R.S.-FNRS) under convention 2.5020.11 and by the Walloon Region.

\bibliographystyle{splncs04}
\bibliography{egbib}

\clearpage
\appendix
\section{Appendix}
\subsection{Preprocessing}
\subsubsection{Imagenet Dataset}
Preprocessing is done by scaling each image to a $224\times224$ resolution and normalizing RGB channels with a mean of [$0.485, 0.456, 0.406$] and a standard deviation of [$0.229, 0.224, 0.225$], as for the training set of ImageNet~\cite{ILSVRC15}.
\subsubsection{CIFAR10 Dataset}
Each image has an original size of $32\times32$ pixels and is normalized using a mean of [$0.4914, 0.4822, 0.4465$] and a standard deviation of [$0.2023, 0.1994, 0.2010$], following the preprocessed values utilized by the ResNet-50 pre-trained model (and the other models) from~\cite{huy_phan_2021_4431043}.
\subsection{Implementation of Saliency Methods}

Gradient-based implementations stem from Captum~\cite{kokhlikyan2020captum}, whereas CAM-based ones come from TorchCAM~\cite{torcham2020}. The implementations of Poly-CAM, RISE as well as CAMERAS come from their authors\footnote{\url{https://github.com/andralex8/polycam}, \url{https://github.com/eclique/RISE}, and \url{https://github.com/VisMIL/CAMERAS} resp.}.
Following common practices, the method parameters are fifty input perturbations (with unit variance Gaussian noise) for SmoothGrad, fifty interpolation steps for Integrated Gradient, four thousand perturbations for RISE, and all the ResNet blocks for Layer-CAM and Poly-CAM. 

\subsubsection{Baseline Replacement}

The baseline replacement with random and uniform comes from the implementation of the random package in the Python Library~\cite{van1995python}. The values for each pixel with random are between 0 (included) and 1 (excluded), while the value chosen for uniform is between the minimum and the maximum in an image or patch.

\subsection{Implementation of Evaluation Metrics}
For faithfulness metrics on ImageNet, the number of steps for each is 224 (in accordance with~\cite{englebert2022polycam}), except for Selectivity, which uses patches. In order to approximately match the number of patches and number of steps, we set the size of these patches to \(14 \times 14\). Similarly, for the CIFAR10 dataset, the patch size is updated to 2 pixels to provide the same ratio (patch width of a fourteenth of the image width). For the Monotonicity Arya metric, we modified the output to return a monotonicity ratio instead of a Boolean result. This is because the metric initially returned only true or false values, which is not comparable to the other metric results. In addition, almost all, if not all of the methods did not increase monotonically over the duration of the metric, so a False value was always returned. This provided little or no information. Providing a percentage allows us to rank the methods on the one hand, and on the other hand to analyze the percentage of monotonicity between the methods, without significantly altering the initial functioning of the metric.

\end{document}